\documentclass[11pt]{article}

\usepackage[margin=1in]{geometry}

\usepackage[english]{babel}
\usepackage{amsmath,amssymb,amsfonts}
\usepackage{dsfont}
\usepackage{changepage}
\usepackage{textcomp,marvosym}

\usepackage[numbers,super,sort&compress]{natbib}
\usepackage{nameref,hyperref}
\usepackage{authblk}

\usepackage[nopatch=eqnum]{microtype}
\DisableLigatures[f]{encoding = *, family = * }

\usepackage[table]{xcolor}
\usepackage{array}
\newcolumntype{+}{!{\vrule width 2pt}}
\newlength\savedwidth

\usepackage{graphicx}
\usepackage{epstopdf}

\usepackage[aboveskip=1pt,labelfont=bf,labelsep=period,justification=raggedright,singlelinecheck=off,font=normalsize,textfont=normalsize]{caption}

\DeclareCaptionFormat{naturecaption}{\textbf{#1#2}#3}
\DeclareMathOperator*{\argmin}{argmin}

\usepackage{lastpage}
\usepackage{fancyhdr}
\begin{document}

% --- REPLACEMENT STARTS HERE ---

\title{\textbf{Emergence of hybrid computational dynamics through reinforcement learning}}

% Use \author and \affil from the authblk package
% The numbers in brackets link authors to their affiliations
\author[1,3]{Roman A. Kononov}
\author[1,2]{Nikita A. Pospelov}
\author[2]{Konstantin V. Anokhin}
\author[1,3]{Vladimir~V.~Nekorkin}
\author[1,3,*]{Oleg V. Maslennikov}

% Define the affiliations
\affil[1]{Institute of Applied Physics of the R.A.S., Nizhny Novgorod, Russia}
\affil[2]{Institute for Advanced Brain Studies of Moscow State University, Moscow, Russia}
\affil[3]{Radiophysics Faculty, Lobachevsky University, Nizhny Novgorod, Russia}
\affil[*]{\textit{Corresponding author:} \href{mailto:olmaov@ipfran.ru}{olmaov@ipfran.ru}}

% Set the date (optional, can be removed to show current date)
\date{}

% This command generates the title block based on the info above
\maketitle

% --- REPLACEMENT ENDS HERE ---

\section*{Abstract}
Understanding how learning algorithms shape the computational strategies that emerge in neural networks remains a fundamental challenge in machine intelligence. While network architectures receive extensive attention, the role of the learning paradigm itself in determining emergent dynamics remains largely unexplored. Here we demonstrate that reinforcement learning (RL) and supervised learning (SL) drive recurrent neural networks (RNNs) toward fundamentally different computational solutions when trained on identical decision-making tasks. Through systematic dynamical systems analysis, we reveal that RL spontaneously discovers hybrid attractor architectures, combining stable fixed-point attractors for decision maintenance with quasi-periodic attractors for flexible evidence integration. This contrasts sharply with SL, which converges almost exclusively to simpler fixed-point-only solutions. We further show that RL sculpts functionally balanced neural populations through a powerful form of implicit regularization—a structural signature that enhances robustness and is conspicuously absent in the more heterogeneous solutions found by SL-trained networks. The prevalence of these complex dynamics in RL is controllably modulated by weight initialization and correlates strongly with performance gains, particularly as task complexity increases. Our results establish the learning algorithm as a primary determinant of emergent computation, revealing how reward-based optimization autonomously discovers sophisticated dynamical mechanisms that are less accessible to direct gradient-based optimization. These findings provide both mechanistic insights into neural computation and actionable principles for designing adaptive AI systems.

\section*{Introduction}
Developing artificial intelligence capable of the flexible, context-dependent decision-making seen in biological systems remains a grand challenge in machine intelligence \cite{botvinick2020deep}. Recurrent neural networks (RNNs) have become central to this pursuit, serving as powerful models for emulating cognitive computations \cite{vyas2020computation, barak2017recurrent} and revealing the deep relationship between network structure and emergent function \cite{durstewitz2023reconstructing}. A key principle underlying both biological and artificial cognition is the use of attractor dynamics, where neural activity settles into stable patterns—such as fixed points for memory maintenance or limit cycles for rhythmic processes—to implement computation \cite{fiete2022attractor, wang2001dynamics}. Recent work has shown that RNNs trained with reinforcement learning (RL) can spontaneously self-organize low-dimensional representations that mirror the functional architecture of the primate brain, establishing attractor dynamics as a convergent strategy for intelligent behaviour \cite{bellafard2024volatile, wang2024working}.

However, a critical gap separates the analysis of fully trained networks from a true understanding of their intelligence: we lack a mechanistic account of how these sophisticated computational structures emerge during the learning process itself. This gap is especially acute for reinforcement learning. While RL's trial-and-error exploration embodies biological learning principles more closely than supervised learning, it remains far less understood from a dynamical systems perspective \cite{song2017reward}. While we know that trained networks employ low-dimensional manifolds \cite{ganguli2023low} and exhibit mixed selectivity \cite{rigotti2013importance}, it remains unclear how the learning algorithm itself—the choice between reward-driven exploration (RL) and explicit error correction (SL)—fundamentally dictates the type of computational solutions that are discovered. This knowledge gap represents a fundamental barrier to creating more robust and adaptive AI, limiting us to empirical success rather than principled design.

Understanding this emergence requires moving beyond analyzing fully trained networks to systematically studying how different learning processes shape computational solutions from the ground up. Previous work has characterized the outcomes of learning—the attractor landscapes and population codes of trained networks—but not the learning process itself as a computational discovery mechanism. Yet RL's reward-driven exploration and SL's explicit gradient descent represent fundamentally different optimization processes that may impose distinct constraints on the space of discoverable solutions. Without direct comparison under controlled conditions, we cannot determine whether these learning paradigms are merely alternative routes to similar solutions or gateways to fundamentally different computational strategies.

Here we establish a framework for understanding how learning algorithms sculpt neural computation. We systematically compare RNNs trained via reinforcement learning and supervised learning on identical context-dependent decision-making tasks \cite{mante2013context}, tracking the emergence of computational solutions from random initialization to expert performance. By analyzing networks throughout training using dynamical systems theory, information theory, and population-level analysis, we reveal whether the learning paradigm itself determines the computational strategies that emerge.

We discover that RL and SL converge on fundamentally different computational solutions, despite solving identical tasks. RL spontaneously discovers hybrid attractor architectures that balance stable fixed-point attractors for decision maintenance with quasi-periodic (oscillatory) attractors for flexible evidence integration. This dynamical richness stands in stark contrast to SL, which overwhelmingly converges on simpler, fixed-point-only solutions. This divergence is mirrored at the population level: RL sculpts functionally balanced neural populations through implicit regularization \cite{turner2023implicit}, while SL produces more heterogeneous and specialized ensembles. The emergence of quasi-periodic dynamics in RL-trained networks is strongly correlated with performance gains and controllably modulated by weight initialization, establishing these complex dynamics as a key mechanism for robust task performance.

By elucidating the evolutionary trajectory of these neural computations, our work provides a bridge between the learning rule and the learned algorithm. We establish that reward-based exploration is not merely an alternative to supervised learning but a fundamentally different optimization process that discovers a richer class of dynamical solutions for temporal processing. These findings offer deep mechanistic insights into how biological circuits may self-organize through reward-driven plasticity \cite{williams1998dopamine, bermudez2021deep} and establish foundational principles for designing next-generation AI systems that can learn flexible and robust behaviors through autonomous interaction with their environment \cite{stroud2024computational, hennig2023emergence}.

% Figure 1: The Setup
\begin{figure}
    \includegraphics{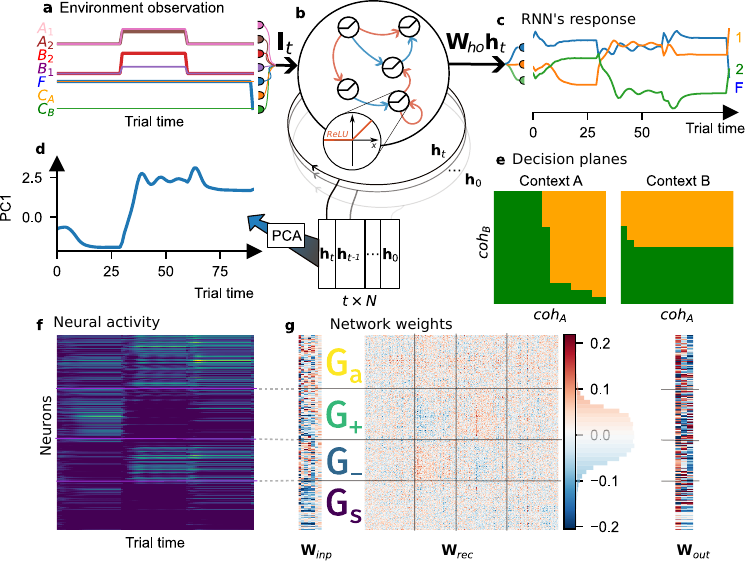}
     \centering
     \caption{
\textbf{Task structure and emergent architecture of a recurrent neural network trained with reinforcement learning.}
(a) Temporal structure of a single trial, showing the distinct fixation, stimulus, delay, and decision stages.
(b) Schematic of the vanilla RNN architecture.
(c) Time-resolved output probabilities from a trained network, illustrating the formation of a decision.
(d) The first principal component of hidden layer activity, capturing the dominant dynamical mode during a trial.
(e) The network's decision boundary in the two-dimensional coherence space, demonstrating successful context-dependent choices.
(f) Raster plot of the full hidden layer activity, revealing complex population-level dynamics.
(g) Recurrent weight matrix of the trained network, reorganized according to functionally distinct neural populations identified in our analysis. The structured connectivity, featuring strong intra-population excitation and inter-population inhibition, was not pre-specified but emerged entirely through reward-driven learning.
}
\label{fig:scheme}
\end{figure}

\section*{Results} \label{sec:Results}

\subsection*{Reinforcement learning discovers a distinct dynamical strategy for decision-making}

To investigate how learning rules shape neural computation, we trained recurrent neural networks (RNNs) on a context-dependent decision-making (CtxDM) task, a canonical benchmark for cognitive flexibility \cite{mante2013context}. In this task, an agent must integrate one of two noisy stimulus streams while ignoring the other, based on a contextual cue. The task unfolds across distinct trial stages—fixation, stimulus presentation, delay, and decision—requiring the network to maintain and manipulate information over time (Fig.~\ref{fig:scheme}a-c). The vanilla RNN architecture, featuring ReLU nonlinearities, was trained using two distinct paradigms: reward-driven Proximal Policy Optimization (PPO), a standard reinforcement learning (RL) algorithm, and gradient-based Adam optimization with a cross-entropy loss, a supervised learning (SL) approach.

\begin{figure}
     \centering
     \includegraphics[width=1.\textwidth]{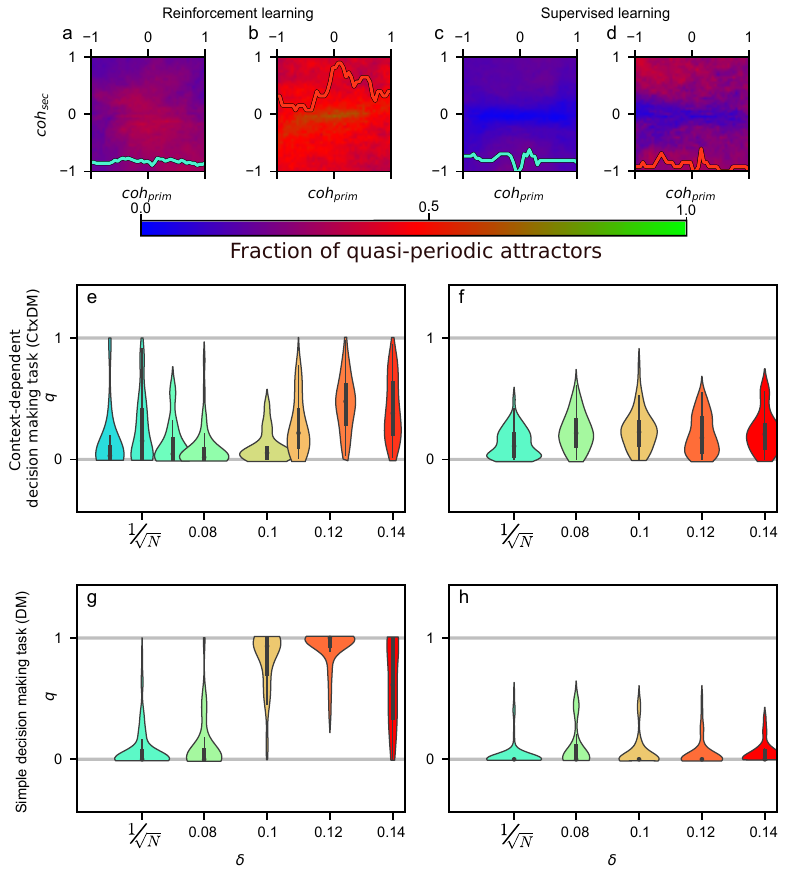}
     \caption{
     \textbf{Reinforcement learning discovers a richer dynamical landscape than supervised learning.}
(a–d) Probability of a stimulus-encoding attractor being quasi-periodic (oscillatory) across the coherence parameter space. The learning algorithm fundamentally alters the solution type: reinforcement learning (a, b) consistently finds oscillatory solutions, particularly for ambiguous, low-coherence stimuli, while supervised learning (c, d) overwhelmingly converges to stable fixed points. This divergence is amplified by broader weight initializations (b, d).
(e–h) Aggregated statistics confirm the prevalence of quasi-periodic attractors is significantly higher in RL than SL across all initialization widths ($\delta$) and for both context-dependent (e, f) and simple (g, h) tasks. These results establish that the learning rule is a primary determinant of the emergent computational strategy, with RL's exploratory nature discovering complex dynamics that SL actively prunes away.
     }
     \label{fig:third}
 \end{figure}

While networks trained under both paradigms can achieve high task accuracy, a deep analysis of their internal dynamics reveals that they converge on fundamentally different computational strategies. We systematically characterized the attractor landscape of trained networks by analyzing the stationary dynamics during prolonged stimulus presentation. This revealed two primary classes of solutions for stimulus encoding: stable fixed points, where neural activity converges to a single static pattern, and quasi-periodic attractors, where activity settles into a persistent, rhythmic oscillation. The choice between these encoding mechanisms represents a foundational difference between the learning paradigms. As shown across large ensembles of networks, RL consistently discovers solutions rich in quasi-periodic dynamics, particularly for ambiguous, low-coherence stimuli. In stark contrast, SL overwhelmingly favors simpler, fixed-point attractors across the entire stimulus space (Fig.~\ref{fig:third}a-d). This divergence suggests that the exploratory nature of RL uncovers a broader and more complex solution space than the explicit error-correction of SL \cite{zhang2021geometric}.

% Figure 2: The Core Finding

This represents a robust and quantifiable signature of the learning rule. The total fraction of quasi-periodic attractors is significantly higher in RL-trained networks across all conditions (Fig.~\ref{fig:third}e-h). Furthermore, this effect is powerfully modulated by the initial weight distribution, a critical hyperparameter in network training. Broader initial weight distributions (a larger half-width $\delta$) dramatically amplify the prevalence of oscillatory dynamics in RL networks, while having a negligible effect on SL networks. This sensitivity to initialization suggests that RL's reward-driven exploration is adept at harnessing initial dynamical richness, whereas SL's supervised gradient descent actively prunes such complexity in favor of the simplest stable solutions \cite{sussillo2013opening}.

This striking divergence in the learned dynamical architectures poses a central question: what is the nature of the computational mechanism that RL discovers, and how does it leverage this richer dynamical repertoire to solve the task? To answer this, we next dissect the evolutionary trajectory of the attractor landscape and the underlying population structures that emerge exclusively during reward-driven learning.

\subsection*{Hybrid attractor dynamics emerge during reward-driven learning}

% Figure 3: Deconstructing the RL Solution
 \begin{figure}
  \centering
  \includegraphics[width=0.75\linewidth]{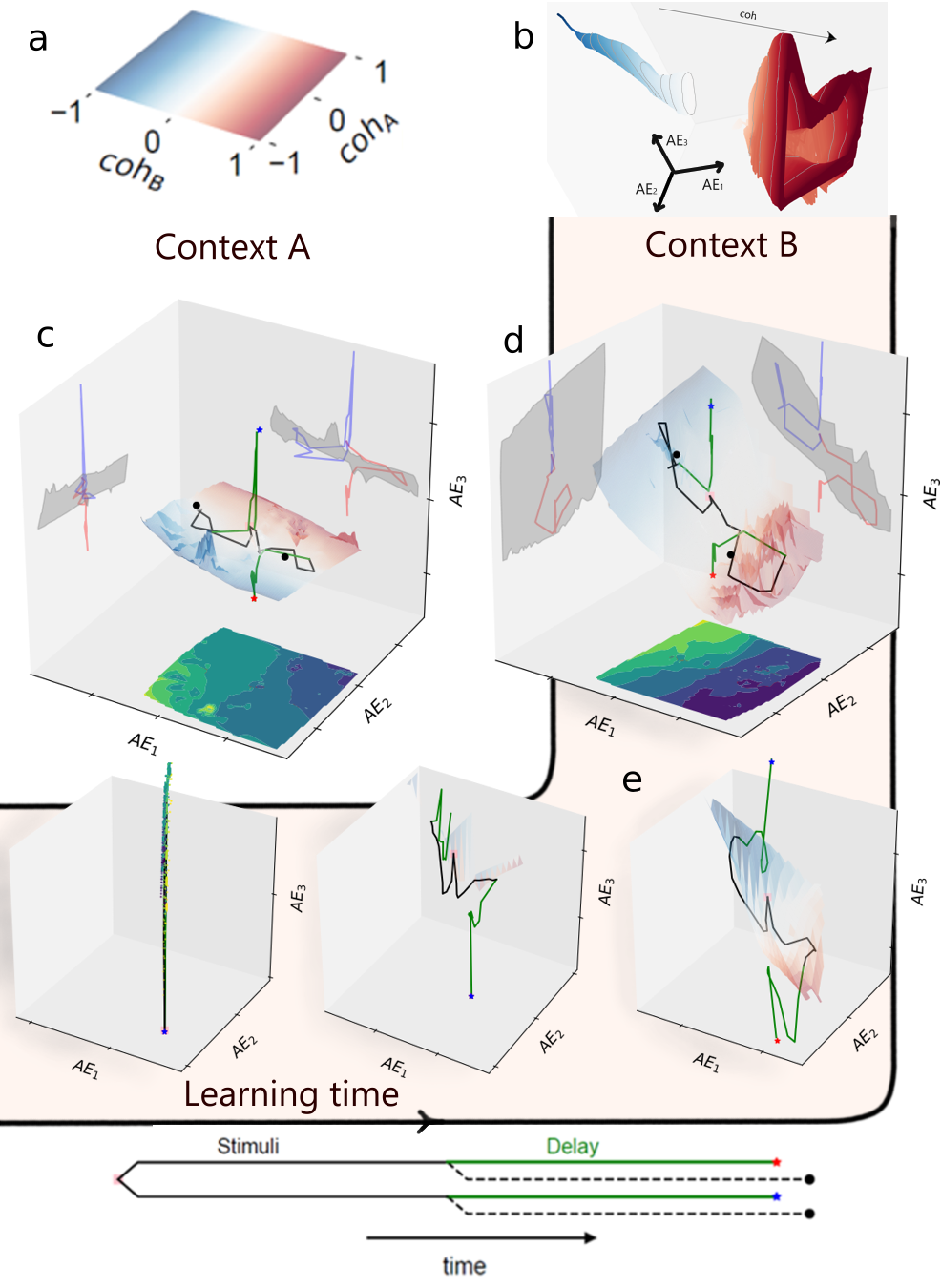}
  \caption{
  \textbf{Emergence of a hybrid attractor architecture during reinforcement learning.}
 % (a)
  (a, b) Low-dimensional projections of neural trajectories for trials with opposite primary coherence, showing distinct paths leading to different choices.
(c–e) Progressive sculpting of the state space across three stages of training. The network learns to separate computational functions into distinct dynamical regimes. Decisions are represented by two discrete, stable fixed-point attractors (blue and red stars), while sensory evidence is encoded along a continuous manifold (colored surface).
An untrained network (c) has an unstructured state space. As training progresses (d), the manifold elongates to separate evidence. The fully trained network (e) exhibits a complete, spontaneous separation of a working memory circuit (the encoding manifold) from a decision circuit (the bistable fixed points), forming a robust hybrid computational system.
}
\label{fig:AE}
\end{figure}

The richer dynamics discovered through RL are not arbitrary but are organized into a sophisticated hybrid computational architecture that elegantly separates stimulus encoding from decision-making. By visualizing the high-dimensional neural trajectories in a low-dimensional space learned by an autoencoder, we can observe the formation of this architecture throughout training (Fig.~\ref{fig:AE}). In the fully trained network, distinct computational functions are mapped onto geometrically distinct regions of the state space. Decisions are represented by two highly stable, discrete fixed-point attractors, providing a robust memory of the categorical choice (Fig.~\ref{fig:AE}e, blue and red stars). In contrast, incoming sensory evidence is encoded along a continuous, low-dimensional manifold. During stimulus presentation, the network state evolves along this surface, its position parametrically encoding the specific combination of primary and secondary coherence values.

This emergent structure constitutes a hybrid dynamical system. It leverages the stability of fixed points for robust, noise-resistant memory of the final choice, while utilizing a more flexible encoding scheme—often employing the quasi-periodic dynamics shown in Fig.~\ref{fig:third} and Supplementary Fig.~\ref{S1_Fig}—for the continuous integration of evidence. During a trial, the network first converges to the encoding manifold (Fig.~\ref{fig:AE}a,b, black trajectories), maintains its state during the delay period (green trajectories), and is finally captured by the basin of attraction of the appropriate decision fixed point. This spontaneous separation of a working memory circuit (the encoding manifold) from a decision circuit (the bistable fixed points) is a hallmark of the RL-derived solution and mirrors functional architectures observed in the primate prefrontal cortex \cite{constantinidis2016role}.

This sophisticated dynamical landscape is not pre-specified but is progressively sculpted by the reward signal during training. In an untrained network, the state space is unstructured; a single, undifferentiated attractor region exists, and trajectories corresponding to different stimulus conditions remain entangled, making the task unsolvable (Fig.~\ref{fig:AE}c). As training progresses, the reward signal guides a gradual stretching and separation of the neural manifold. In a partially trained network, the stimulus-encoding surface begins to elongate along the primary coherence axis, creating distinct paths for positive and negative evidence, even before the two decision attractors have fully bifurcated and separated (Fig.~\ref{fig:AE}d). This intermediate stage highlights how performance gains are directly coupled to the geometric refinement of the network's state space. The final, fully trained architecture (Fig.~\ref{fig:AE}e) represents the culmination of this process, where an optimal separation of encoding and decision subspaces has been achieved.

This emergent separation of computational roles into distinct dynamical regimes—stable fixed points for choice and a flexible manifold for evidence—is a defining feature of the RL solution. We next investigate how this functional architecture is physically implemented at the level of neural populations.

\subsection*{Population-level self-organization implements the emergent dynamics}

% Figure 4: The Population-level Implementation
\begin{figure}
\centering
\includegraphics[width=0.95\textwidth]{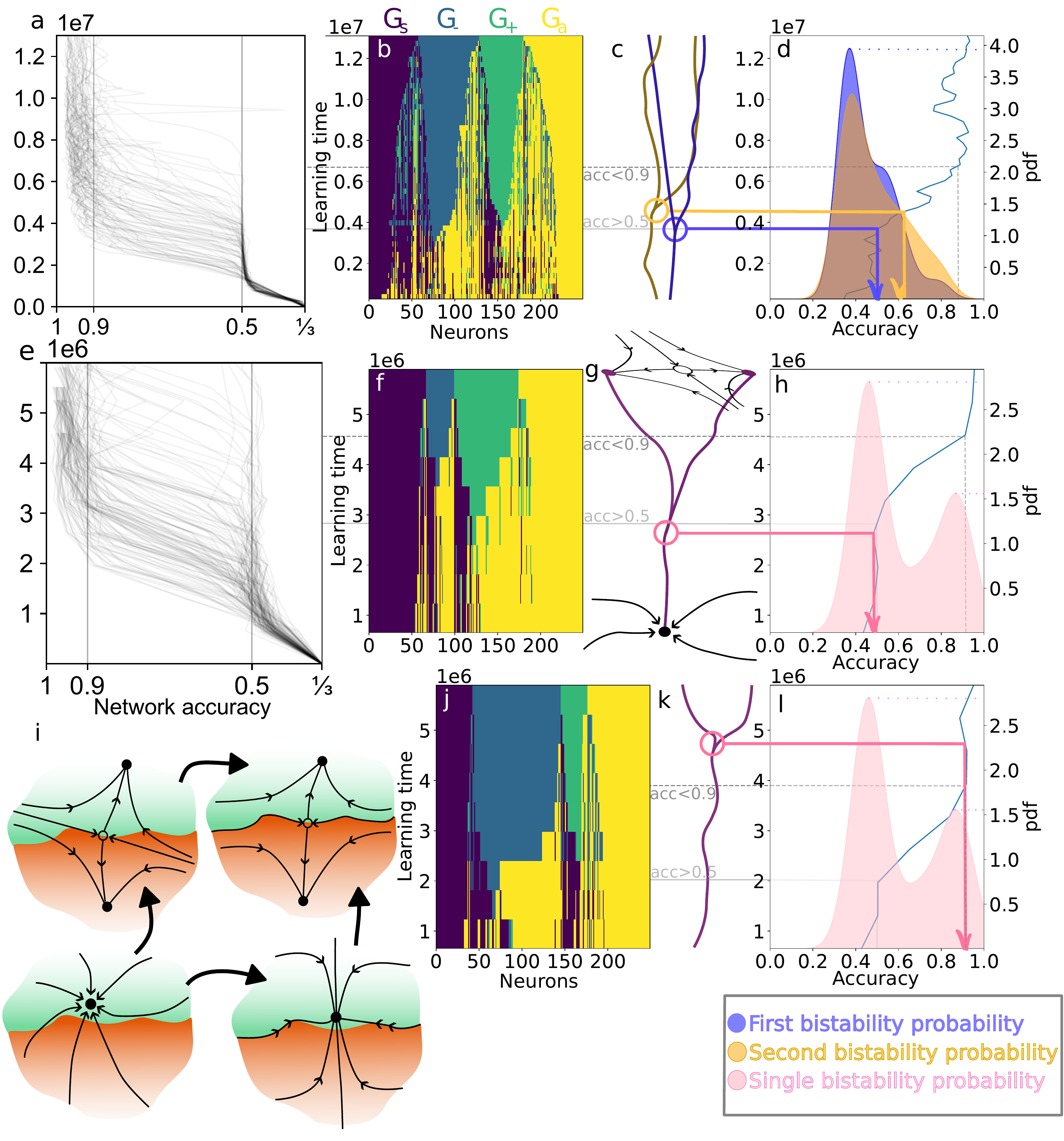}
\caption{
\textbf{Population self-organization is coupled to performance gains and bistability.}
(a, e) Learning curves showing accuracy improvements during training.
(b, f, j) Four functionally distinct neural populations emerge during RL, defined by their selective participation in attractor dynamics. Notably, the opposing coherence-selective populations ($\mathrm{G}_+, \mathrm{G}_-$) are sculpted to be approximately equal in size.
(c, g, k) The emergence of a bistable decision landscape, allowing for two distinct choices.
(d, h, l) Statistical analysis reveals that the jump in task accuracy from the 0.5 plateau is tightly correlated with both the formation of specialized populations and the emergence of decision bistability. This demonstrates a direct link between the network's structural organization, its dynamical capabilities, and its functional performance.
(i) A schematic illustrating the bifurcation process leading to a bistable decision landscape.
}
\label{fig:second}
\end{figure}

The hybrid dynamical architecture discovered through reinforcement learning is not an abstract property but is physically implemented by the self-organization of the network into functionally specialized and structurally balanced neural populations. By analyzing neuronal activity during prolonged stimulus presentation across the coherence space, we identified four distinct populations based on their participation in the attractor dynamics: a silent population ($\mathrm{G}_s$), two populations selectively active for positive ($\mathrm{G}_+$) or negative ($\mathrm{G}_-$) primary coherence, and a population active for any non-zero coherence ($\mathrm{G}_a$). The emergence and interaction of these groups provide the substrate for the network's computational capabilities.

The formation of this population structure is inextricably linked to the learning process and performance improvements. We tracked the evolution of these populations alongside task accuracy and the emergence of decision bistability (Fig.~\ref{fig:second}). Early in training, when performance is at chance level, the network lacks distinct decision attractors and specialized populations. The critical leap in performance, escaping the 0.5 accuracy plateau, coincides precisely with two key events: the bifurcation of the decision dynamics into a bistable system and the emergence of the two opposing, coherence-selective populations, $\mathrm{G}_+$ and $\mathrm{G}_-$ (Fig.~\ref{fig:second}b,c,d). 
A defining characteristic of the RL solution is that these two opposing populations are sculpted to be approximately equal in size. This population balance is not incidental; it represents a robust structural motif discovered by RL. By ensuring that competing representations for positive and negative evidence are supported by equally-sized neural ensembles, the network avoids over-specialization and maintains flexibility—a hallmark of effective regularization.
This population balance stands in stark contrast to SL-trained networks, which typically converge to highly heterogeneous and imbalanced population structures despite achieving similar task accuracy (see Supplementary Fig.~\ref{S2_Fig}), and thus represents a specific and robust outcome of the implicit regularization imposed by reward-driven exploration.

% Figure 5: Functional Validation of Populations
\begin{figure}
     \centering
     \includegraphics[width=0.9\textwidth]{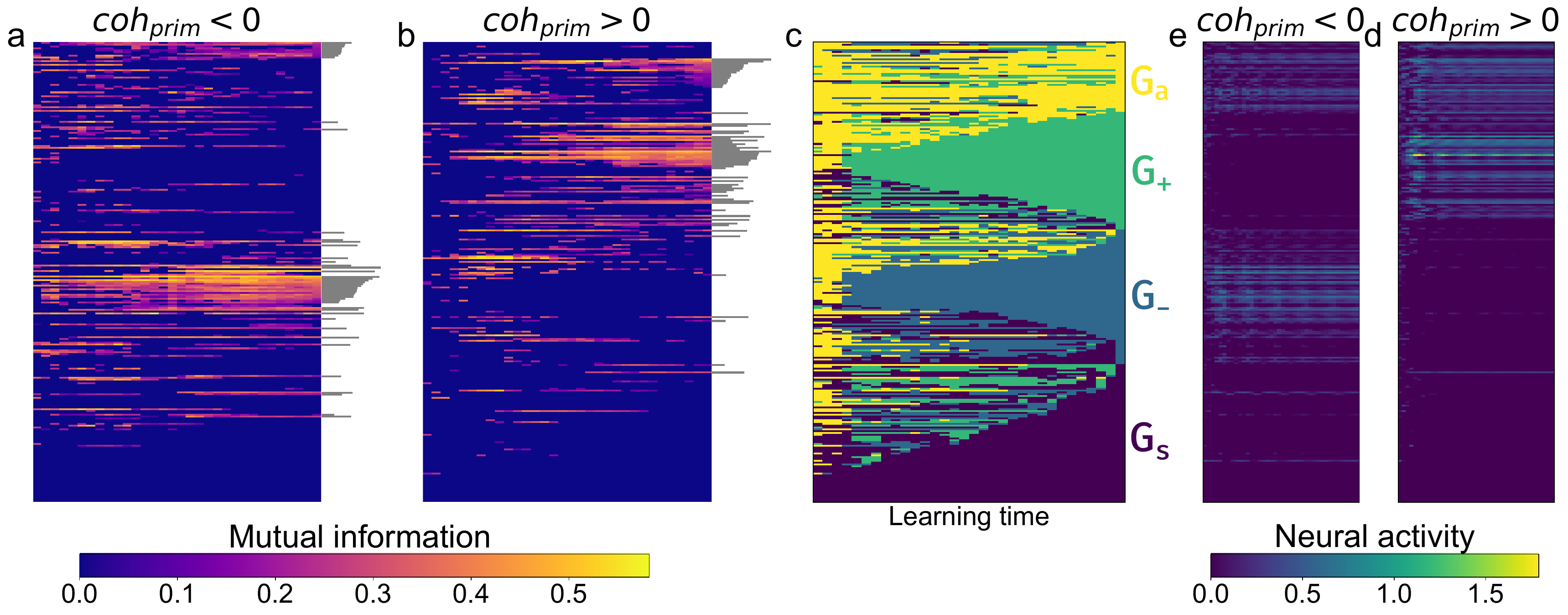}
     \caption{
          \textbf{Dynamically-defined populations exhibit distinct information encoding roles.}
(a, b) Mutual information (MI) between single-neuron activity and the sign of primary coherence. Neurons are ordered according to the population structure identified in Fig.~\ref{fig:second} (shown in panel c).
(c) The four functional populations ($\mathrm{G}_s, \mathrm{G}_+, \mathrm{G}_-, \mathrm{G}_a$) identified via dynamical systems analysis.
(d, e) Example single-neuron activity traces demonstrating selective responses.
The MI analysis validates the functional roles of the populations: neurons in $\mathrm{G}_+$ selectively encode positive coherence, while those in $\mathrm{G}_-$ encode negative coherence. This reveals a clear division of labor and confirms that the emergent population structure forms the basis for reliable information processing.
     }
     \label{fig:MI}
\end{figure}

Information-theoretic analysis validates that these dynamically-defined populations have distinct functional roles in encoding task variables. Mutual information (MI) calculations confirm that neurons within the $\mathrm{G}_+$ population selectively encode information about positive primary coherence, while neurons in $\mathrm{G}_-$ encode negative coherence (Fig.~\ref{fig:MI}). This analysis reveals a fine-grained functional organization, demonstrating a clear division of labor that aligns perfectly with the population structure identified through dynamical systems analysis. Intriguingly, it also reveals substructure within the context-integrating population ($\mathrm{G}_a$), which contains non-overlapping subgroups of neurons with opposing stimulus preferences, highlighting a sophisticated mechanism for context-dependent routing of information.

Finally, this functional organization is imprinted upon the network's anatomical connectivity. Reorganizing the recurrent weight matrix according to population membership reveals a clear and intuitive circuit structure (Fig.~\ref{fig:scheme}g). We observe strong positive recurrent connections (excitation) among neurons within the same coherence-selective population ($\mathrm{G}_+$ to $\mathrm{G}_+$, $\mathrm{G}_-$ to $\mathrm{G}_-$) and strong negative connections (inhibition) between the opposing populations ($\mathrm{G}_+$ to $\mathrm{G}_-$). This is the classic connectivity motif of a winner-take-all circuit, which enables robust decision-making. Crucially, this effective architecture was not designed by hand but emerged purely through reinforcement learning. Thus, the exploratory process of RL forges not just abstract computational dynamics, but a physically structured, self-organized circuit that robustly implements those dynamics.

\section*{Discussion}
Our study establishes a fundamental principle for understanding intelligent systems: the learning algorithm itself is a primary determinant of the emergent computational strategy. By systematically dissecting how recurrent neural networks (RNNs) develop solutions for context-dependent decision-making, we reveal that reinforcement learning (RL) and supervised learning (SL) sculpt fundamentally different dynamical landscapes. The central finding is that RL-trained networks spontaneously discover a hybrid computational scheme, leveraging both stable fixed-point attractors for robust memory maintenance and quasi-periodic attractors for flexible temporal integration. This demonstrates that the nonlinear dynamics crucial for advanced computation \cite{maass2002perturbations,maslennikov2022nonlinear} are not just a property of the final network, but a direct outcome of the learning process. This emergent dynamical richness, where the network's capabilities are intrinsically linked to the geometry of its state-space organization \cite{yang2020artificial}, highlights RL's capacity to discover solutions that naturally balance persistence with adaptability.

The stark divergence between learning paradigms provides a mechanistic explanation for their differing capabilities. While SL-trained networks, guided by explicit error signals, predominantly converge to simpler, fixed-point-dominated solutions \cite{sussillo2013opening, zhang2021geometric}, RL agents operate under a different imperative. By optimizing for long-term reward through trial-and-error exploration \cite{mohan2023exploration}, RL is incentivized to discover more complex and robust computational mechanisms. The higher prevalence of quasi-periodic dynamics in RL networks is not an incidental byproduct but a functional adaptation, enabling multi-timescale processing and evidence integration within a single, unified architecture. This discovery suggests that for tasks requiring dynamic integration of ambiguous information, RL's exploratory nature acts as an asset, discovering hybrid analog-digital computation strategies that are difficult to engineer explicitly \cite{pan2021hybrid}. Furthermore, we find that RL promotes the emergence of functionally balanced neural populations, which act as a powerful form of implicit regularization \cite{turner2023implicit}, preventing over-specialization and enhancing generalizability.

While our work establishes a strong correlation between the learning paradigm and the emergent dynamical strategy, it also poses a critical question of causality. Are the quasi-periodic dynamics functionally essential, or are they an elegant but non-essential byproduct—a computational spandrel—of reward-based exploration? We propose the former. The systematic emergence of these dynamics, particularly for ambiguous inputs where integration is most critical, and their strong correlation with performance gains, suggests a definitive functional role. Future work involving targeted causal interventions, such as optogenetically-inspired manipulations to stabilize or disrupt these oscillations in silico, could definitively test their contribution to robust decision-making.

This discovery of a dual-attractor system provides a compelling computational model that bridges a long-standing debate in neuroscience regarding the nature of working memory. While persistent, stable activity has been considered the hallmark of memory maintenance \cite{pmc2023decoding, pmc2023population, fuster1971neuron}, emerging evidence advocates for dynamic coding mechanisms that rely on evolving patterns of activity \cite{pnas2023optimal}. Our findings suggest this is not an either-or scenario. Instead, reward-driven learning naturally produces circuits that exploit both strategies: fixed points to encode categorical decisions and quasi-periodic dynamics to handle the ambiguity and temporal dependencies inherent in sensory evidence. Rather than modeling specific synaptic rules, our work provides a high-level computational framework for understanding the functional outcome of dopamine-dependent plasticity mechanisms in the prefrontal cortex, which are known to shape both stable representations and flexible updates \cite{williams1998dopamine}.

The temporal evolution of these mechanisms during training offers a new window into how complex representations are formed. We observed a prolonged period of refinement where attractor landscapes stabilize long after behavioral performance has plateaued. This process of "crystallization" \cite{bellafard2024volatile} strongly parallels the "grokking" phenomenon in machine learning, where generalization suddenly emerges after a long period of apparent memorization \cite{power2022grokking}. This suggests a potentially general principle of learning in complex systems: achieving robust, noise-resistant internal models requires extended consolidation that is not always visible in immediate performance metrics.

The functional realism of our RL-trained models extends to multiple levels of organization. The emergent push-pull connectivity between functionally distinct neural populations mirrors the microcircuit organization of the primate prefrontal cortex \cite{constantinidis2016role}. The spontaneous development of neurons with mixed selectivity for various task parameters, without any explicit instruction, recapitulates a key feature of higher-order cortical areas \cite{rigotti2013importance} and aligns with findings from activation maximization studies showing that neurons often respond to complex feature mixtures \cite{ponce2019evolving, neurotrain}. Crucially, these biologically-plausible features emerge not from a hand-designed model, but as a consequence of applying a general learning principle—reward maximization—to a flexible substrate~\cite{pugavko2023multitask}. This supports the view that many hallmark features of cortical computation may be convergent solutions discovered through goal-directed learning.

These insights translate directly into principles for designing more capable and robust AI. First, the discovery of quasi-periodic attractors provides a blueprint for building systems that can process information across multiple timescales without explicit clocks, a feature seen in coupled attractor models of brain function \cite{compte2025dynamic}. Second, the balance between regularization and expressivity achieved through population-level dynamics offers a new strategy for developing reliable cognitive architectures \cite{gilbert2024balancing}. Specifically, our results suggest that hybrid training schemes—using SL to rapidly establish a coarse task solution and then using RL to fine-tune and enrich the network's dynamical repertoire—could be a powerful method for developing more adaptive agents. These principles have profound implications for neuromorphic computing \cite{sainath2019reinforcement,ivanov2025neural} and could inform the development of more efficient learning algorithms guided by topological and dynamical priors \cite{carlsson2020topological}.

Several limitations merit consideration. First, our analysis focused on a single task paradigm; extending this framework to other cognitive domains will be essential for assessing generalizability. Second, the vanilla RNN architecture, while interpretable, lacks the biological realism of spiking networks or models with explicit Dale's law constraints. Third, our comparisons were limited to PPO and Adam optimizers; examining other RL algorithms and supervised methods could reveal additional nuances in the learning-dynamics relationship.

In conclusion, our work reframes the role of reinforcement learning from a mere training tool to a discovery engine for novel computational principles. By demonstrating how reward-driven exploration autonomously discovers sophisticated, multi-modal dynamical strategies, we bridge deep learning theory with systems neuroscience \cite{richards2019deep}. The principles unveiled here—concerning the emergence of hybrid attractors, population-level regularization, and prolonged representational refinement—are foundational and likely to be broadly applicable across diverse domains, from robotics to natural language processing \cite{wang2022parametric}. Ultimately, a deep, mechanistic understanding of the learning process itself is the key to moving beyond engineering solutions by imitation and toward designing truly intelligent systems from first principles.

\section*{Methods}

\subsection*{RNN Model and Task Specification}
All numerical experiments utilized a vanilla recurrent neural network (RNN) defined by the state update equation:
 \begin{equation}
     \mathbf{h}_{t+1} = \text{ReLU}(\mathbf{W}_{hh}  \mathbf{h}_t + \mathbf{W}_{ih} \mathbf{I}_t),
     \label{eq:RNN}
 \end{equation}
where $\mathbf{h}_t \in \mathbb{R}^{N_{hidden}}$ is the hidden state vector at time $t$. 
The network size of $N_{hidden} = 250$ was chosen to be consistent with similar studies of RNNs on cognitive tasks. This size is sufficiently large to allow for the emergence of complex dynamics and population structures, yet small enough to remain computationally tractable for training large ensembles of networks. $\mathbf{I}_t \in \mathbb{R}^{N_{input}}$ is the input vector ($N_{input} = 7$), $\mathbf{W}_{hh}$ is the recurrent weight matrix, and $\mathbf{W}_{ih}$ is the input weight matrix. The agent's action policy is derived from the hidden state via a linear readout $\mathbf{o}_{t} = \mathbf{W}_{ho} \mathbf{h}_t$, followed by a softmax function to produce action probabilities.

The input vector $\mathbf{I}_t$ for each trial was composed of a noise-free signal $\mathbf{I}^{pure}$ and an additive noise term:
\begin{equation}
     \mathbf{I}_t = \mathbf{I}^{pure}_t + \alpha \mathbf{r}_t,
\end{equation}
where $\mathbf{r}_t$ is a random vector with components drawn from $\mathcal{U}(-1,1)$. The noise-free input $\mathbf{I}^{pure}$ encodes the seven task variables: a fixation signal  $(F)$, two pairs of stimulus inputs corresponding to the two contexts ($A_1$, $A_2$ and $B_1$, $B_2$), and two context cues ($C_A$, $C_B$).

Stimulus coherence for each context was defined as:
 $$coh_A = A_1 - A_2, \quad coh_B = B_1 - B_2.$$
The trial structure consisted of four stages: fixation, stimuli, delay, and decision, each with a specific input configuration.

Network weights were initialized from a uniform distribution: recurrent and input weight matrices from $\mathcal{U}(-\delta, \delta)$, and readout weights from $\mathcal{U}(-0.1, 0.1)$. We systematically varied the initialization width $\delta$ across experiments to examine its effect on emergent dynamics, as detailed in the Results.

\begin{table}[!ht]
\centering
\caption{Hyperparameter configuration for reinforcement and supervised learning algorithms used in ensemble training.}\label{tab:hyper}
\begin{tabular}{ |p{3cm}|p{1cm}||p{3cm}|p{1cm}|  }

 \hline
 \multicolumn{2}{|c||}{PPO} & \multicolumn{2}{|c|}{Adam}\\
 \hline
 $\gamma$              & 0.99   & $\beta_1$ & 0.9\\
 $\lambda_{gae}$         & 0.95   &  $\beta_2$ & 0.999 \\
 $\epsilon$& 0.2    &  learning rate  & 0.0003 \\
 entropy coef.           & 0      &        max grad norm&  0.5\\
 value func. coef.           & 0.5    &        &  \\
\hline
\end{tabular}

\end{table}

\subsection*{Network Training Procedures}
We trained four ensembles of networks across the two tasks (DM and CtxDM) and two learning paradigms. Training was terminated when a network achieved 95\% accuracy.

\subsubsection*{Reinforcement Learning}
Networks were trained using the Proximal Policy Optimization (PPO) algorithm, an actor-critic method implemented in Stable-Baselines3. Hyperparameters were set according to Table~\ref{tab:hyper} to ensure stable learning.

\subsubsection*{Supervised Learning}
Supervised baselines were trained using the Adam optimization algorithm with a cross-entropy loss function. The target was "fixation" for all timesteps except the final decision step. See Table~\ref{tab:hyper} for hyperparameters.

\subsubsection*{Computational Details}
All simulations were performed using Python 3.9, PyTorch 1.12, and Stable-Baselines3 1.6. Ensemble sizes and weight initialization widths ($\delta$) varied by experiment as detailed in the Results. Hyperparameters listed in Table~\ref{tab:hyper} were selected based on preliminary experiments and standard practices for each algorithm.

\subsection*{Dynamical Systems and Population Analysis}
\subsubsection*{Attractor Identification and Classification}
To identify attractors, we extended the duration of the stimulus stage and allowed the system dynamics to converge to a stationary state. For a given set of active neurons, the dynamics are governed by a linear system $\mathbf{h}^{act}_{t+1} = \mathbf{W}_{act} \mathbf{h}^{act}_t + \mathbf{s}_{act}$. An attractor was classified as a stable fixed point if the leading eigenvalue $\mu_{lead}$ of the Jacobian matrix $\mathbf{W}_{act}$ satisfied $|\mu_{lead}| < 1$. It was classified as quasi-periodic if the leading complex conjugate eigenvalue pair satisfied $|\mu_{con}^{lead}| > 1$.

\subsubsection*{Population Definition}
We classified neurons into functional groups using a template-matching procedure based on their complete response profiles across the coherence space. This objective method provides a robust definition of functional specialization.

\begin{figure}
    \includegraphics[width=0.75\linewidth]{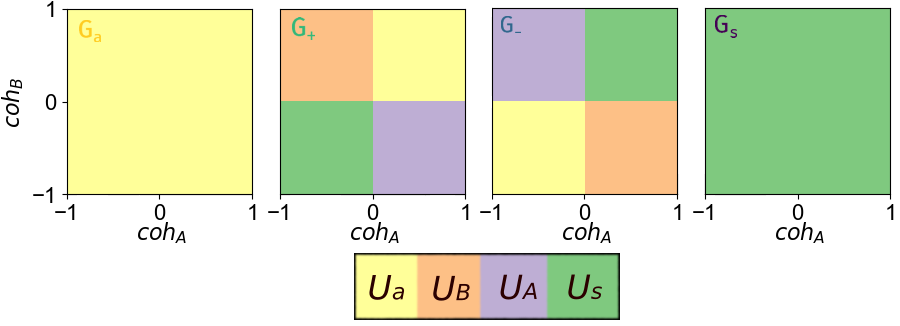}
    \caption{\textbf{Idealized templates used for neural population classification.}
    Idealized templates used for neural population classification. The four panels show the canonical activation patterns that define the functionally distinct neural populations. Each matrix represents a single neuron’s theoretical activation profile across the two-dimensional coherence space. Crucially, the entries are not binary but are integer labels representing context-specific participation in stationary dynamics (e.g., 0 for silent, 1 for active in Context A only, 2 for active in Context B only, 3 for active in both contexts). These templates correspond to the four primary populations that consistently emerge in trained networks: $\mathrm{G}_a$ (active across contexts), $\mathrm{G}_+$ (selective for positive primary coherence), $\mathrm{G}_-$ (selective for negative primary coherence), and $\mathrm{G}_s$ (silent). In our analysis, each neuron from a simulation was computationally assigned to the population whose template, $\mathbf{M}^c$, provided the best least-squares fit to its observed activity matrix, $\mathbf{M}$ (see Methods for the formal definition).
    }
    \label{fig:evo_labs}
\end{figure}

First, for each neuron, we constructed an activity-type matrix $\mathbf{M}$ over a $44 \times 44$ grid of coherence values ($coh_A$ vs. $coh_B$). For each point in this grid, we determined the neuron’s participation in the stationary dynamics for each of the two possible contexts (Context $A$ and Context $B$). Based on this, we assigned one of four integer labels to the corresponding entry in the neuron’s matrix $\mathbf{M}$:
\begin{itemize}
\item 0 ($U_s$): The neuron is silent (does not participate) in the stationary dynamics for either context.
\item 1 ($U_A$): The neuron participates exclusively in the dynamics for Context A.
\item 2 ($U_B$): The neuron participates exclusively in the dynamics for Context B.
\item 3 ($U_a$): The neuron participates in the dynamics for both contexts.
\end{itemize}
Second, we defined four idealized template matrices, $\mathbf{M}^c$ (Fig.~\ref{fig:evo_labs}), which represent the canonical activity-type profiles for the populations that consistently emerged in trained networks ($\mathrm{G}_s, \mathrm{G}_+, \mathrm{G}_-, \mathrm{G}_a$). Finally, each neuron was assigned to the population $c$ whose template $\mathbf{M}^c$ minimized the squared error with its observed activity-type matrix $\mathbf{M}$:
$$\argmin_{c \in (\mathrm{G}_s,\mathrm{G}_+,\mathrm{G}_-,\mathrm{G}_a)} \left(\sum_{i=1}^{N_g} \sum_{j=1}^{N_g} (M_{i,j}-M_{i,j}^c)^2 \right).$$

\subsection*{Information-Theoretic Analysis}
Neuronal selectivity was quantified using Mutual Information (MI) between single-neuron activity and task variables. We used the computationally efficient Gaussian Copula Mutual Information (GCMI) method \cite{gcmi2016statistical}. Statistical significance was assessed through permutation testing: for each neuron, we generated a null distribution from 10,000 temporally shuffled surrogates and compared the observed MI against this distribution. Multiple comparisons were controlled using Holm's sequential procedure with family-wise error rate $\alpha = 0.01$.

\subsection*{Dimensionality Reduction and Visualization}
To visualize high-dimensional neural trajectories, we trained a three-layer autoencoder with a bottleneck size of $N_{code}=3$. To ensure interpretable and consistent visualizations across training epochs, we added an alignment term to the standard $L_2$ reconstruction loss. This alignment term encourages the primary axes of the latent space to consistently correspond to the primary task variables ($coh_{prim}$ and $coh_{sec}$), enabling meaningful comparison of state-space geometry across different stages of training. The full loss function was:
\begin{equation}
    \mathcal{L} = ||A -\tilde{A}||_F^2 + \lambda \mathcal{L}_{align},
\end{equation}
where $\mathcal{L}_{align} = - (R(C_1, coh_{prim}) + R(C_2, coh_{sec}))$, with $R$ being the Pearson correlation coefficient, $C_{1,2}$ being the first two latent dimensions, and $\lambda=0.1$.

\section*{Author Contributions}
R.A.K. and O.V.M. designed the study. R.A.K. and N.A.P. performed the computational experiments, and analyzed the data. K.V.A., V.V.N., and O.V.M. supervised the research. N.A.P. and O.V.M. wrote the manuscript. All authors reviewed and approved the final version.

\section*{Competing Interests}
The authors declare no competing interests.

\section*{Data Availability}
All data generated during this study, including trained network models, experimental configurations, dynamical analysis results, and population classification data, are available from the corresponding author upon reasonable request.

\section*{Code Availability}
The MI-based analysis of neuronal selectivity, as well as dimensionality reduction, was performed using the DRIADA package \cite{driada2025software}. All other code is available from the corresponding author upon reasonable request.

\section*{Acknowledgments}  \label{sec:Acknowledgments}
This research was supported by the Russian Science Foundation (projects 23-72-10088 and 24-12-00245) and the Non-Commercial Foundation for Support of Science and Education ``INTELLECT''.

\newpage
\section*{Supporting information}
\setcounter{figure}{0}
\renewcommand{\thefigure}{S\arabic{figure}}

% Supplementary Figure S1
\begin{figure}[htbp]
    \centering
    \includegraphics[width=0.7\linewidth]{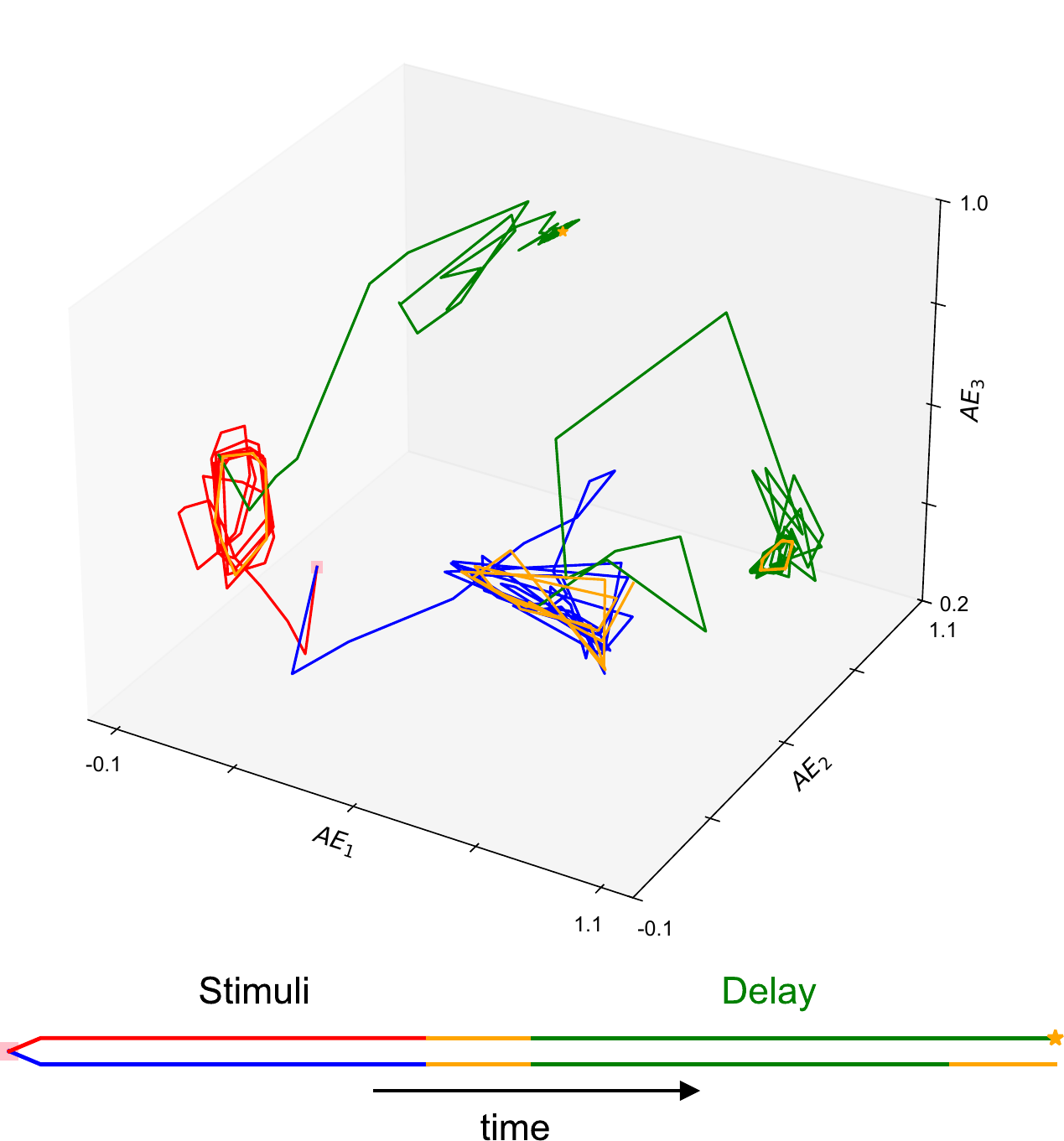}
    \caption{\textbf{Example of a quasi-periodic stimulus-encoding attractor.} Neural trajectories are shown during a prolonged stimulus presentation for trials with opposite coherence signs. Instead of converging to a single fixed point, the network state settles into a stable, rhythmic oscillation (orange curves), demonstrating a dynamic mechanism for maintaining stimulus information. This provides a concrete example of the oscillatory dynamics whose prevalence is quantified in Fig.~\ref{fig:third}.}
    \label{S1_Fig}
\end{figure}

% Supplementary Figure S2
\begin{figure}[htbp]
    \centering
    \includegraphics[width=0.8\linewidth]{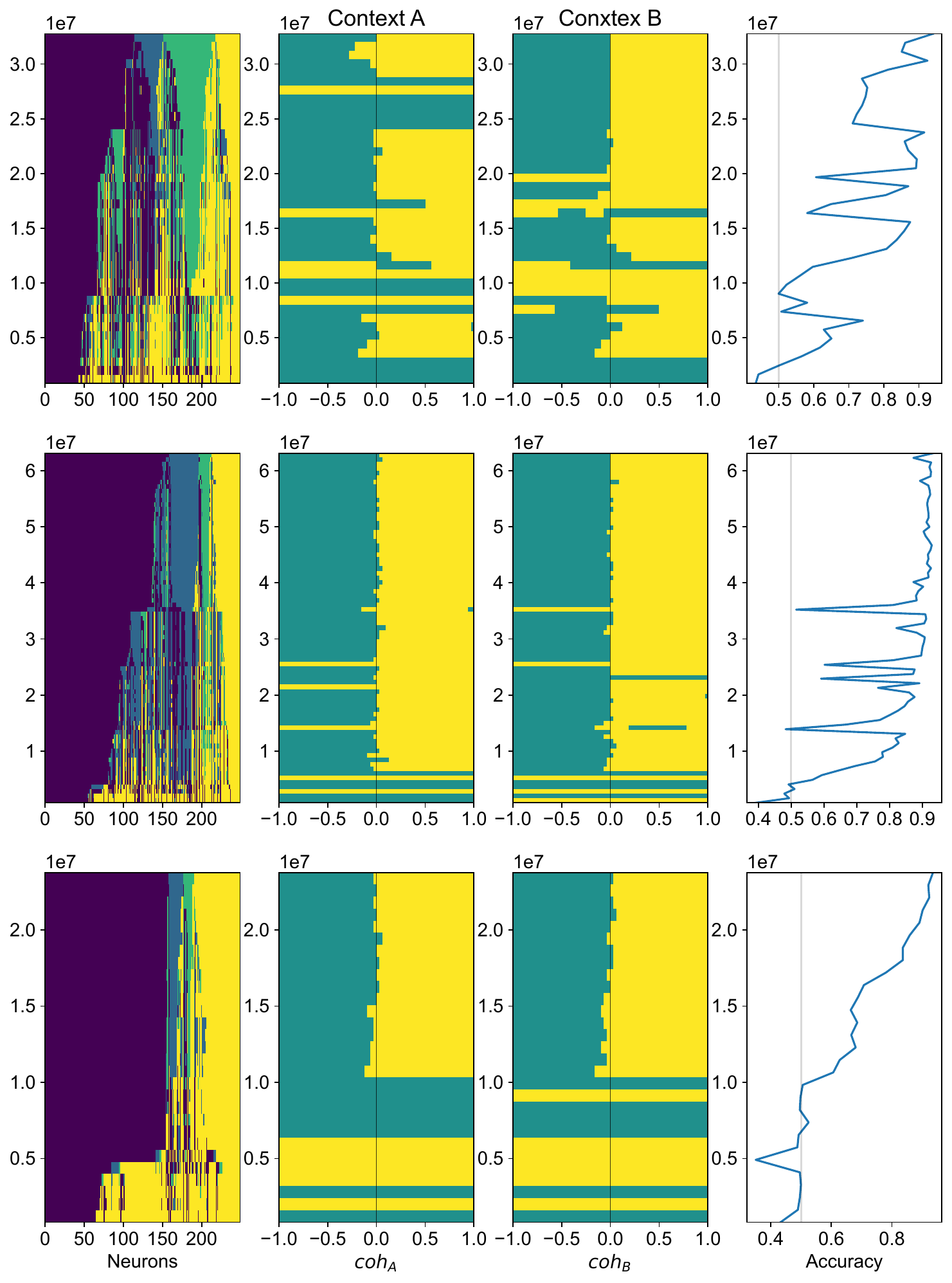}
    \caption{\textbf{Heterogeneous population structure in networks trained with supervised learning.} Analogous analysis to Fig.~\ref{fig:second} for SL-trained networks. While SL can also produce bistable decision dynamics, the underlying coherence-selective populations that emerge are significantly more heterogeneous and imbalanced in size compared to the balanced architectures consistently found through reinforcement learning. This highlights a key structural difference between the solutions discovered by the two paradigms.}
    \label{S2_Fig}
\end{figure}

% Supplementary Figure S3 (formerly S4)
\begin{figure}[htbp]
    \centering
    \includegraphics[width=0.8\linewidth]{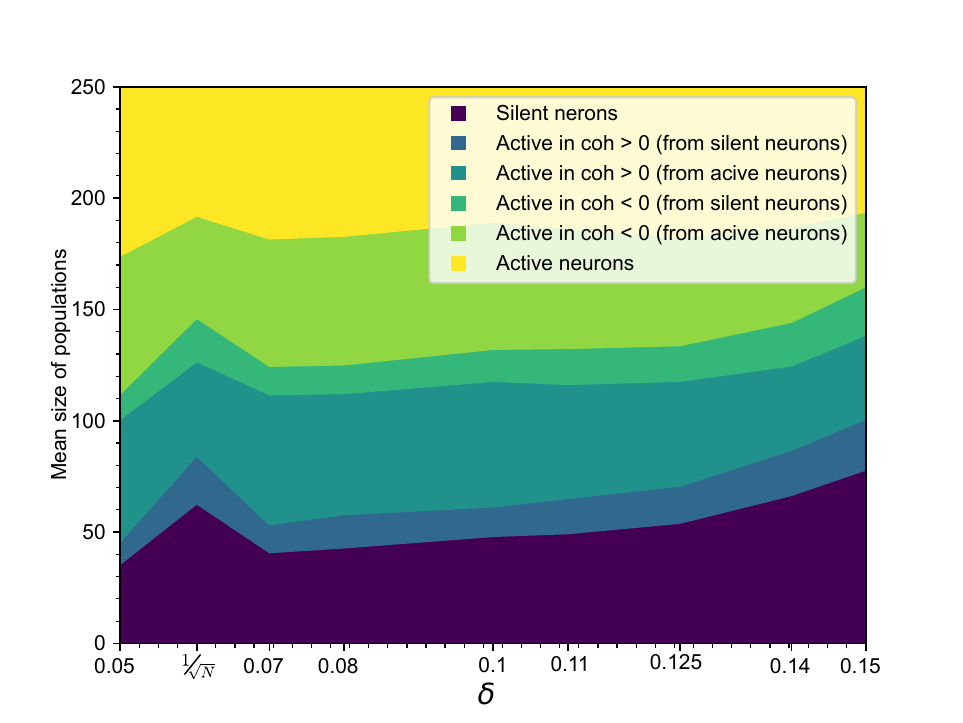}
    \caption{\textbf{Weight initialization width modulates population structure in RL networks.} Mean sizes of functionally defined populations as a function of the initialization width $\delta$. Broader initializations, which promote the quasi-periodic dynamics seen in Fig.~\ref{fig:third}, are associated with a larger silent population ($\mathrm{G}_s$). This suggests a mechanistic link where RL leverages a pool of uncommitted neurons to construct more complex, oscillatory dynamics when the initial state of the network is more dynamically rich.}
    \label{S4_Fig}
\end{figure}

% Supplementary Figure S4 (formerly S5)
\begin{figure}[htbp]
    \centering
    \includegraphics[width=1\linewidth]{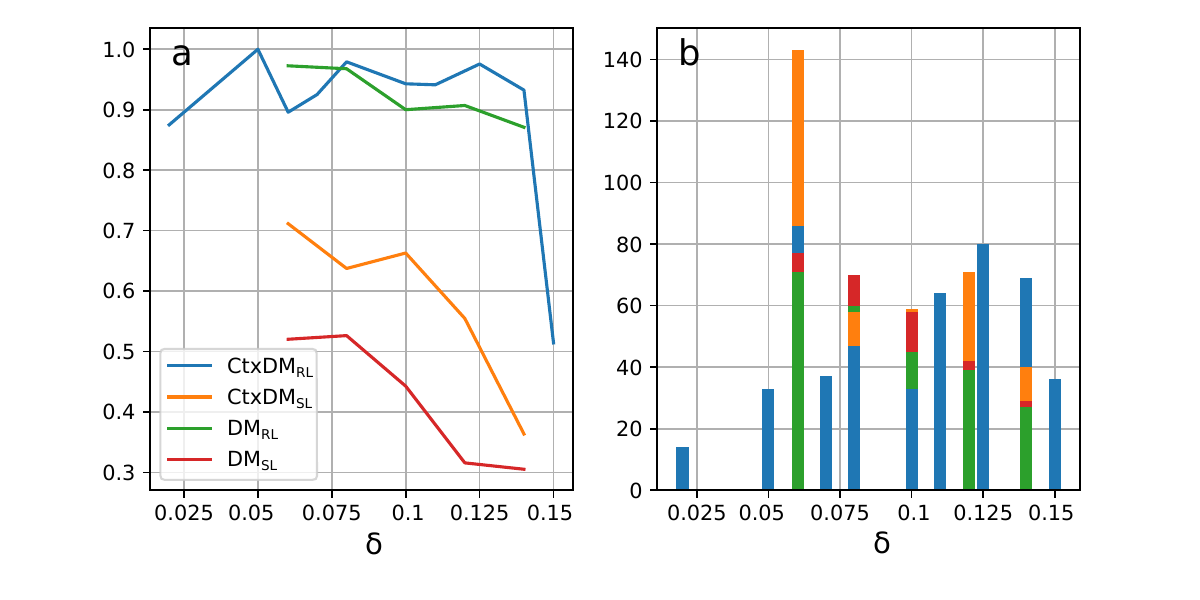}
    \caption{\textbf{Training stability and final ensemble sizes.}
(a) Probability of successful network training as a function of weight initialization width $\delta$. Very narrow or very wide initializations can lead to training failures (e.g., gradient explosion/vanishing).
(b) The number of successfully trained networks in each experimental ensemble used for the statistical analyses presented in the main text.
}
    \label{S5_Fig}
\end{figure}

% Supplementary Figure S5 (formerly S6)
\begin{figure}[htbp]
    \centering
    \includegraphics[width=1\linewidth]{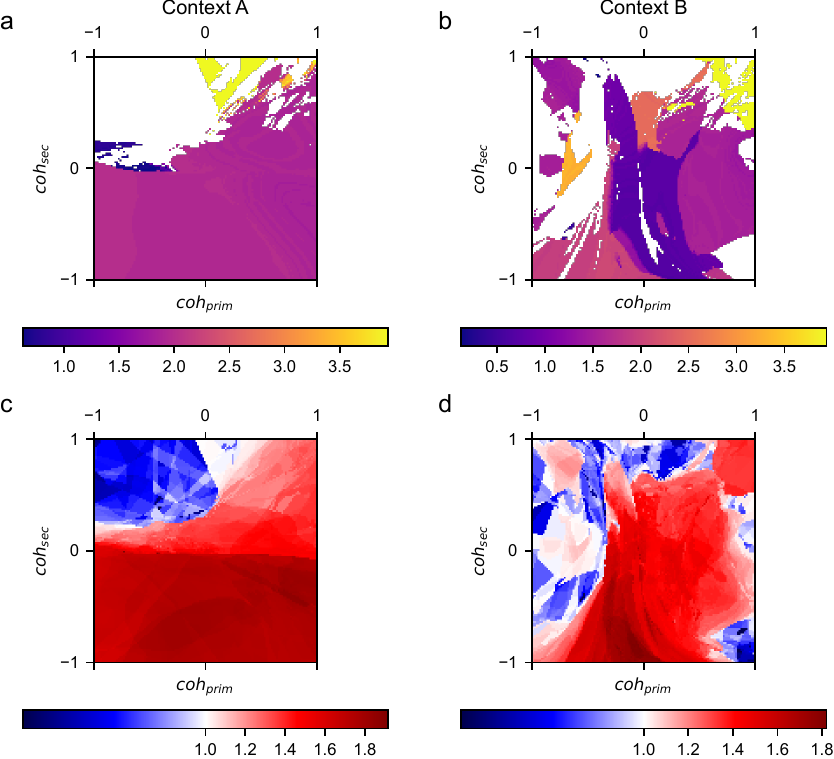}
    \caption{
    \textbf{Linear stability analysis confirms attractor classification.}
(a-b) Distribution of dominant oscillation frequencies for quasi-periodic attractors across the coherence space for a representative RL-trained network. White regions correspond to stable fixed-point attractors.
(c-d) The modulus of the leading eigenvalue of the system's Jacobian matrix for each corresponding attractor. Values less than 1 (blue/green) indicate a stable fixed point, while values greater than 1 (red) indicate an instability that gives rise to an oscillatory (quasi-periodic) attractor. The spatial alignment between panels (a-b) and (c-d) validates the method used to classify attractor types throughout the paper.
}
    \label{S6_Fig}
\end{figure}

\end{document}